\pdfoutput=1
\documentclass{article} %
\usepackage{iclr2027_conference,times}

\usepackage{amsmath,amsfonts,bm}

\def\eqref#1{equation~\ref{#1}}
\def\1{\bm{1}}

\DeclareMathAlphabet{\mathsfit}{\encodingdefault}{\sfdefault}{m}{sl}
\SetMathAlphabet{\mathsfit}{bold}{\encodingdefault}{\sfdefault}{bx}{n}

\usepackage[hidelinks]{hyperref}
\usepackage{url}
\usepackage{graphicx}
\usepackage{booktabs}
\usepackage{amsmath,amssymb,amsthm}
\usepackage{xcolor}
\usepackage{placeins}
\usepackage{subcaption}

\graphicspath{{figures/}}

\newcommand{\maybefig}[2]{\includegraphics[width=#2]{figures/#1}}

\title{Are Coreset Selection Methods Worth Their Cost?}

\author{Yangze Liu \\
Shandong University \\
\texttt{yangze2@illinois.edu}
\And
Zhongyi Han\thanks{Corresponding author.} \\
Shandong University \\
\texttt{zhongyi.han@sdu.edu.cn}}

\iclrfinalcopy
\begin{document}

\maketitle
\lhead{Preprint.}

\begin{abstract}
Data selection methods, often called coreset selection, pick a training subset to make training cheaper. The usual evaluation reports downstream accuracy at a fixed subset size, which leaves two quantities out of the metric: the time spent selecting the subset, and the training recipe behind each reported number. We close that gap with an end-to-end benchmark that standardizes downstream training and prices selection and training on one wall-clock axis estimated from controlled timing probes, spanning 4 datasets from CIFAR-10 to ImageNet-1K, 11 selectors, 5 fractions, and 3 seeds, across 1{,}674 released runs with their selected indices. Repeated-sampling work has shown that budget-aware evaluation already favors random strategies, and our two budget studies test whether that holds when every selector is granted its most favorable tested operating point. Across eight budget anchors on each of CIFAR-10 and Tiny ImageNet, no anchor is won by a sophisticated selector: every winner is class-balanced random sampling, repeated random sampling, or full-data training. In nearby-cost comparisons on ImageNet-1K, training on all data for fewer epochs has the highest accuracy among the operating points we probe. A per-dataset cost audit shows that selection cost is dominated at every scale by a fixed full-dataset scan, so it cannot be amortized away by selecting a smaller fraction, and its absolute size does not extrapolate from one dataset to another. We further separate cost amortization at a fixed architecture from cross-architecture accuracy transfer, and document 9 correctness fixes to a widely used codebase, one of which shifts a standard Herding baseline by $5.9$ percentage points (pp). Selection time is not free preprocessing, and an evaluation that ignores it measures the wrong quantity.
\end{abstract}

\section{Introduction}
\label{sec:introduction}

Data has replaced compute as the scarce input in large-scale training. Frontier language models are trained on curated fractions of the web rather than on all of it, and agent pipelines now generate synthetic trajectories faster than they can be trained on, so deciding which examples deserve gradient steps is no longer optional \citep{sorscher2022beyond,marion2023less,albalak2024survey}. Every such pipeline reduces to one primitive: a selector scores the candidate pool, a fixed subset is kept, and a model is trained on that subset. Scoring the pool is itself a training-scale computation, since the selector must run a proxy model, extract features, or compare pairs over the whole pool before it can discard anything. The claim that selection saves training cost therefore has to be tested with the selector's own cost on the bill. This paper tests that claim for the select-then-train primitive, usually called coreset selection, in the supervised vision setting where the entire pipeline can be timed end to end under one recipe.\footnote{We measure the primitive where an end-to-end protocol is feasible today. LLM pretraining data selection, dynamic data pruning, dataset distillation, and active learning change the evaluation primitive (token-level scoring, data reuse, or label acquisition), and extending the cost accounting to them is future work.} Published selectors span random sampling, geometry-based matching (Herding \citep{welling2009herding}), training-dynamics scoring (Forgetting \citep{toneva2019forgetting}, GraNd/EL2N \citep{paul2021deep}), and optimization-based selectors (CRAIG \citep{mirzasoleiman2020craig}, GradMatch \citep{killamsetty2021gradmatch}). Most evaluations compare equally sized subsets on downstream accuracy alone. The question a practitioner faces is which strategy produces the best trained model under a fixed wall-clock budget once the time spent selecting the subset is counted.

Existing protocols are hard to act on because two practical determinants are usually uncontrolled. First, selection cost is usually left out of the comparison, even though every non-trivial method we audit touches the entire training set, via probe training, feature extraction, or pairwise computation, before it can discard any sample. This upfront cost can dominate the budget in the small-subset regimes where selection is meant to save the most training time. Second, downstream accuracies are not directly comparable across papers. On ImageNet-1K the full-data reference point itself is not shared: Dataset Quantization \citep{zhou2023data} uses ResNet-18 and MoSo \citep{tan2023moso} uses ResNet-50, while CCS and $\mathbb{D}^2$ Pruning \citep{zheng2023coverage,maharana2024d2} both use ResNet-34 and $\mathbb{D}^2$ states that it follows CCS's hyperparameters, yet their full-data anchors disagree by $0.44$\,pp ($73.54$ against $73.1$). Reference points differ across backbones and disagree even under a shared recipe, so gains over baseline cannot be placed on a common axis.

\begin{figure}[!ht]
  \centering
  \maybefig{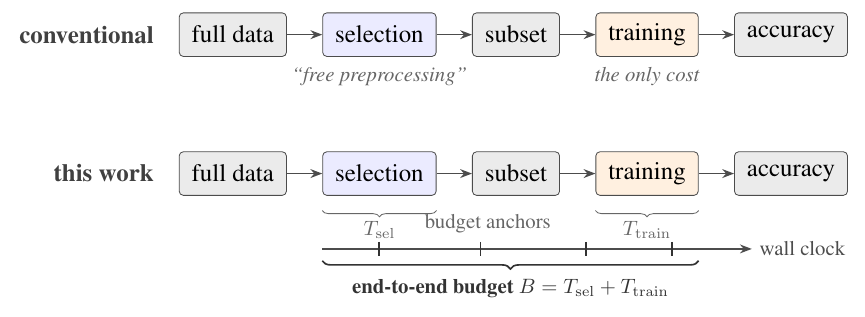}{0.72\linewidth}
  \caption{End-to-end selection protocol: the benchmark prices selection cost and downstream training on one wall-clock axis under a unified recipe, then compares strategies at matched budgets.}
  \label{fig:pipeline}
\end{figure}

We build an end-to-end benchmark (Fig.~\ref{fig:pipeline}) that standardizes downstream training and prices selection and training on the same wall-clock axis, which changes the unit of comparison from subset size to consumed training opportunity: a selector is useful only if its accuracy gain exceeds what the same cost could buy by training on more uniformly sampled data. The main grid spans 4 datasets from 6K to 1.28M samples, with additional studies on a medical-imaging domain and alternative training protocols.

We push the budget question further than a fixed operating-point comparison. RS2 \citep{okanovic2024rs2} showed that repeated random sampling beats standard one-shot selectors under training-time budgets, and its without-replacement variant is equivalent to full-data training for fewer epochs under an adapted learning-rate schedule. Our additional question is where selection spends the training opportunity that random strategies can use immediately. We audit the fraction-independent cost floor on each dataset and let every method choose its most favorable tested (fraction, epochs) operating point under the same downstream recipe, which separates a cost that smaller subsets cannot remove from a cost whose absolute size must be measured again on each dataset (Section~\ref{sec:tradeoff}).

\paragraph{Contributions.} The benchmark protocol prices selection and downstream training on one wall-clock axis under a unified recipe, and scores every method at its own most favorable tested operating point inside a matched budget. The release carries corrected DeepCore-based code with 9 documented fixes, the timing-audit tools, raw and processed result CSVs, and the selected indices from 1{,}674 runs, so the grids can be reanalyzed without re-running a selector. Sections~\ref{sec:budget-grid} to~\ref{sec:amortization} report the budget grids, the ImageNet nearby-cost comparison, the accuracy landscape, the cost decomposition, and the reuse analysis.

\section{Related Work}
\label{sec:related-work}

\paragraph{Budget-aware training and repeated sampling.}
Budgeted training studies how to schedule optimization when the epoch budget is fixed in advance \citep{li2020budgeted}. Closest to our headline question, RS2 \citep{okanovic2024rs2} showed that repeatedly drawing a fresh random subset each epoch reduces time-to-accuracy and outperforms standard one-shot selectors under training-time budgets. Its without-replacement construction traverses a shuffled full dataset and is equivalent to full-data training for fewer epochs, and RS2 also adapts the learning-rate schedule to the reduced number of updates. We include RS2 as a first-class strategy and treat compressed full-data schedules as prior art. Our budget grids extend that comparison by granting each selector its most favorable operating point under one recipe and one measured cost axis. We also localize the cost to a fixed full-dataset scan (Section~\ref{sec:tradeoff}), document implementation corrections (Appendix~\ref{app:bugfixes}), and separate cost amortization from accuracy transfer (Section~\ref{sec:amortization}).

\paragraph{Pruning quality and cheaper selection.}
Selection via Proxy reduces selection overhead using smaller or shorter-trained proxy models \citep{coleman2020selection}, and scalable self-supervised scores offer another route \citep{sorscher2022beyond}. CCS stratifies importance scores to preserve coverage at high pruning rates \citep{zheng2023coverage}, while TDDS aggregates training dynamics beyond a single snapshot \citep{zhang2024tdds}. These approaches move the cost and quality tradeoff along axes our implemented baselines do not cover, and dynamic methods such as InfoBatch prune and reweight samples during training \citep{qin2024infobatch}.

\paragraph{Prior benchmarks.}
The two closest prior efforts to ours are \emph{Deep Learning on a Data Diet} \citep{paul2021deep} and \emph{DeepCore} \citep{guo2022deepcore}. Data Diet compares training-dynamics scoring rules (GraNd, EL2N, Forgetting) on CIFAR-scale data but reports no selection cost and covers neither ImageNet-1K nor geometry- or optimization-based methods. DeepCore re-implements 12 methods across CIFAR-10 and ImageNet-1K under a unified library, but reports accuracy-vs-fraction results only, without clean end-to-end timing. We additionally identify 9 correctness bugs in DeepCore's public codebase (including an argmax/argmin sign inversion in Herding that alone depresses CIFAR-10 accuracy at $f{=}0.1$ by ${\sim}5.9$\,pp, Appendix~\ref{app:bugfixes}), so numerical conclusions based on the original implementation should be rechecked. Two more recent benchmarks are orthogonal to ours: \emph{DRoP} \citep{vysogorets2025drop} evaluates pruning through worst-class robustness, and \citet{dharmasiri2025coreset} study bias and worst-group accuracy on spurious-correlation benchmarks. None of these works places \emph{selection cost} and \emph{training cost} on the same axis under one single-machine protocol: Data Diet, DeepCore, and DRoP report accuracy at fixed fractions and treat selection as free. Our audit puts $95$ to $99\%$ of selection time in a fixed full-dataset scan (Section~\ref{sec:tradeoff}), so a fixed-fraction accuracy carries no cost information.

\section{Benchmark Design}
\label{sec:design}

\paragraph{Datasets.}
We evaluate on 4 primary datasets (Table~\ref{tab:datasets}) covering natural-image classification across scales, with CUB-200 as a fine-grained setting. Additional results on Retinal OCT (83K medical images, 4 classes) in Appendix~\ref{app:retinaloct} test whether conclusions transfer beyond natural images.

\begin{table}[!ht]
  \caption{Dataset summary and full-data baseline accuracy (ResNet-18 trained from scratch, mean over 3 seeds).}
  \label{tab:datasets}
  \centering
  \begin{tabular}{lcccc}
    \toprule
    Dataset & \#Train & \#Classes & Image size & Full-data acc.\ (\%) \\
    \midrule
    CIFAR-10      & 50{,}000      & 10   & $32\times32$   & $95.42$ \\
    Tiny ImageNet & 100{,}000     & 200  & $64\times64$   & $66.65$ \\
    CUB-200       & 5{,}994       & 200  & $224\times224$ & $49.17$ \\
    ImageNet-1K   & 1{,}281{,}167 & 1000 & $224\times224$ & $67.82$ \\
    \bottomrule
  \end{tabular}
\end{table}

\paragraph{Methods.}
We evaluate 11 methods spanning five families distinguished by their selection-time computation: Random, Uniform (class-balanced random), Herding, Submodular (GraphCut $+$ LazyGreedy \citep{minoux1978accelerated}), Craig, Forgetting, GraNd, EL2N, Moderate \citep{xia2023moderate}, class-balanced greedy k-center (kCenter), following \citet{sener2018active}, and GradMatch. EL2N, Moderate, and kCenter were added under the same interface (Appendix~\ref{app:new-methods}). All methods except Random run in class-balanced mode. Craig, GradMatch, and Submodular are excluded on ImageNet because their $O(\sum_k n_k^2)$ complexity is infeasible at 1.28M samples. Throughout the paper we call a strategy \emph{simple} if it can begin downstream training without first scanning the full dataset or training a probe model (Random, Uniform, RS2, and full-data training), and \emph{sophisticated} otherwise. The Coverage-centric Coreset Selection algorithm of \citet{zheng2023coverage} is a different method and is listed in Appendix~\ref{app:scope}.

\paragraph{Scope of the main grid.}
The main grid covers fixed-subset \emph{select-then-train} methods only. Appendix~\ref{app:scope} distinguishes compatible but unevaluated methods, including Selection via Proxy and the original CCS, from methods with different training or data primitives. RS2 \citep{okanovic2024rs2}, a per-epoch resampling strategy rather than a fixed-subset selector, enters the budget studies (E1) as a strategy baseline because the budget axis is where it is designed to compete.

\paragraph{Selection-signal protocol.}
Random and Uniform require no selector model. All learned selection signals come from a from-scratch ResNet-18 trained within the benchmark on the same split, never from ImageNet-pretrained or external proxy embeddings, since ImageNet-1K is itself in the benchmark and pretrained features would conflate selection quality with transfer-learning quality. The selector trains for 40 epochs on CIFAR-10, CUB-200, Retinal OCT, and Tiny ImageNet, and 10 epochs on ImageNet-1K. Our GraNd and EL2N results use an explicit \emph{epoch-1 variant}: scores are averaged over 10 independently initialized one-epoch probes, and method labels refer to this variant throughout, which limits probe-training work while retaining initialization averaging. Herding, Moderate, and kCenter use last-layer embeddings from this selector model. kCenter starts from the point nearest the class centroid and then applies farthest-first traversal within each class. CRAIG, GradMatch, and GraphCut Submodular use gradient-derived objectives, and Forgetting accumulates forgetting events during selector training. All selector training, feature extraction, scoring, and greedy or optimization steps count toward selection time, after which the selector model is discarded and downstream training starts from a fresh ResNet-18 initialization.

\paragraph{Training protocol.}
Downstream models use ResNet-18 trained from scratch with dataset-specific hyperparameters fixed across methods, so per-method tuning cannot confound the comparison: SGD with momentum 0.9, Nesterov acceleration, weight decay $5\times10^{-4}$, cosine decay, and fixed crop-and-flip augmentation. Canonical training lasts 200 epochs except on ImageNet-1K (90). Appendix~\ref{app:hparams-pointer} specifies batch sizes, learning rates, and augmentations. Fixed-fraction analyses use peak test accuracy, and E1 and E2 use recorded final accuracy. E1 retains a sophisticated canonical cell's peak as a favorable bound when its final is missing, and excludes missing-final simple cells (Appendix~\ref{app:final-acc}). Scores average 3 seeds unless flagged otherwise. Separate ResNet-50 transfer and ViT-Tiny studies appear in Appendices~\ref{app:e5-table} and~\ref{app:robustness}.

\paragraph{Budget grids (E1).}
On CIFAR-10 and Tiny ImageNet, fractions $f\in\{0.05,0.1,0.3,0.5,0.7,1.0\}$ are crossed with epoch counts matching $fE\approx\{10,20,60,100\}$ full-data epochs, capped at 200 epochs, and CIFAR-10 additionally includes schedules up to 400 epochs at the two smallest fractions. Every method's canonical 200-epoch cell also competes. Eight anchors span each grid: four cheapest Uniform costs at the compute tiers above, plus the 10th, 50th, and 90th cost percentiles and the maximum. Each method chooses its highest three-seed mean score among affordable cells, with selection charged at its audited cost. Craig, GradMatch, and Submodular receive zero selection cost as a favorable bound because the audit does not cover them (Appendix~\ref{app:e1-full}).

\paragraph{ImageNet nearby-cost duels (E2).}
Two search groups, approximately 10 and 18 GPU-hours, compare compressed full-data training, Uniform, and audited sophisticated selectors at reallocated operating points (Table~\ref{tab:e2-decomp}). Searches are single-seed, with the higher group's podium cells and lower group's top two cells repeated over 3 seeds. Reduced-epoch schedules compress cosine decay to the target length \citep{li2020budgeted}. These compressed full-data points are the strategy equivalent to RS2 without replacement \citep{okanovic2024rs2}, listed as Full.

\paragraph{Timing methodology.}
The benchmark's training runs were executed across several hosts, GPU types, and load levels, all on GPUs of one architecture generation so that no heterogeneous hardware enters the accuracy comparison. Their wall clocks record scheduling history as much as the work a method requires, so a cost axis built from them would not be comparable across methods. We therefore price every (method, dataset, fraction) cell with a dedicated timing probe on one idle RTX 4090 host, under one driver and one software stack, with fixed CPU affinity, worker count, and thread limits, so every cost in this paper is comparable by construction. Selection is timed directly, and training is estimated from 2 warmup epochs, 10 timed epochs, and 3 timed test passes on a Uniform subset, retaining the audit's evaluation-rate convention when scaling to other epoch counts (Appendix~\ref{app:timing-probe}). The audit covers ImageNet, Tiny ImageNet, and CUB-200, with a later extension to CIFAR-10, and E1, E2, and the R18 amortization study price both selection and training from it. The probed fractions are $\{0.05,0.1,0.3,0.7\}$, with $f{=}0.5$ interpolated and $f{=}1$ linearly extrapolated, checked against a direct full-data probe on the same host and settings that agrees within $+0.40$ to $+1.79\%$ on the four datasets and leaves every budget conclusion unchanged (Appendix~\ref{app:full-probe-check}). Sensitivity analyses perturb these probe estimates rather than substituting run clocks.

\paragraph{Artifact scope.}
Appendix~\ref{app:artifacts} lists the released code, audit and analysis scripts, per-run result files, and selected indices, recorded against the canonical splits without redistributing raw images.

\section{Results}
\label{sec:results}

\subsection{Budget grids: no anchor goes to sophisticated selection}
\label{sec:budget-grid}

\begin{figure}[!ht]
  \centering
  \begin{subfigure}[t]{0.48\linewidth}
    \centering
    \maybefig{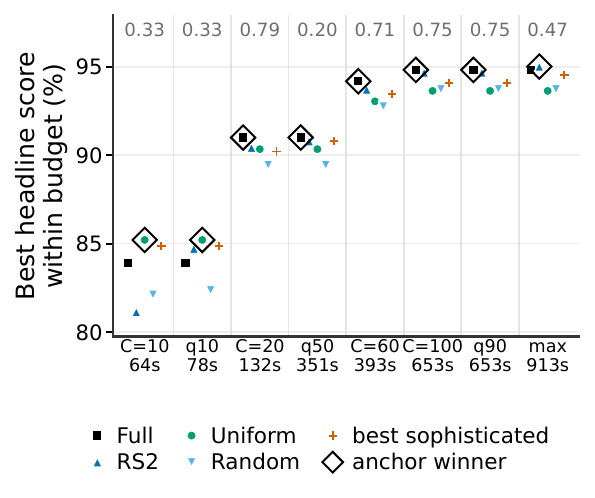}{\linewidth}
    \caption{CIFAR-10 budget anchors.}
    \label{fig:budget-grid}
  \end{subfigure}\hfill
  \begin{subfigure}[t]{0.48\linewidth}
    \centering
    \maybefig{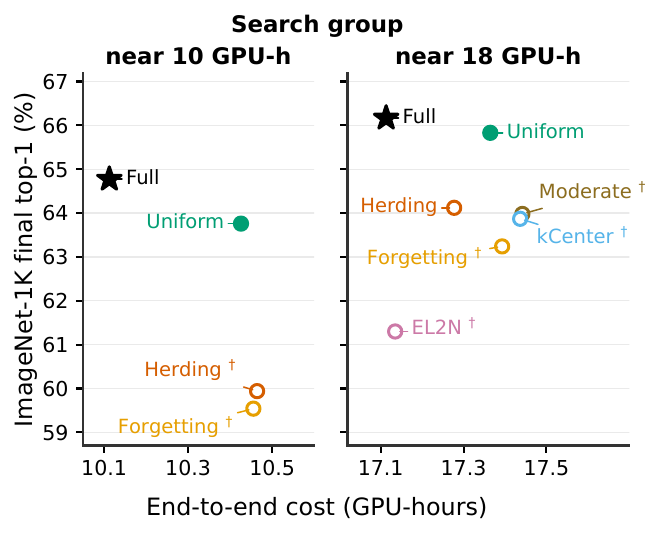}{\linewidth}
    \caption{ImageNet-1K nearby-cost duels.}
    \label{fig:duel}
  \end{subfigure}
  \caption{Budget accounting on two testbeds. (a) Best three-seed mean
  headline score reachable within each of the eight CIFAR-10 anchors, with
  the anchor winner ringed and the winner minus best-sophisticated gap in pp
  printed above each group. (b) ImageNet-1K operating
  points in the two search groups, each on its own probe-estimated
  end-to-end cost axis so that nearby points separate, with daggers marking
  single-seed points.}
  \label{fig:budget-row}
\end{figure}

Selection cost settles the cheapest anchors before accuracy enters. Figure~\ref{fig:budget-grid} shows, at each of the eight CIFAR-10 anchors, the best three-seed mean headline score each strategy reaches within that budget over all (fraction, epochs) operating points with selection cost included, with sophisticated selectors pooled into one entry. Every audited non-trivial selector has a \emph{selection floor}: it becomes affordable only once the budget covers the probe training and full-dataset scan that selection requires. On CIFAR-10 that floor is roughly $84$ to $241$ audited seconds against a cheapest grid cell of $63$ seconds, and on Tiny ImageNet it rises to about $10$ to $31$ minutes (Fig.~\ref{fig:budget-grid-tiny}). At the two lowest anchors no audited selector is affordable, so the pooled entry there is Submodular at its zero-cost bound (Section~\ref{sec:design}), while class-balanced random sampling is already training. At anchors the sophisticated selectors can afford, their best operating points still do not beat the best simple strategy at the same cost. The cheap anchors are settled by the floor and the rest by accuracy at matched cost.

Every one of the sixteen anchors goes to a simple strategy. On CIFAR-10 the winners are Uniform at the two cheapest anchors, full-data training at the next five, and RS2 at the maximum anchor. Tiny ImageNet also yields eight simple-strategy winners, RS2 or full-data training at every anchor (Tables~\ref{tab:e1-cifar} and~\ref{tab:e1-tiny}). Appendix~\ref{app:e1-full} gives operating points, runner-up margins, and the seed and cost sensitivity of each decision, and Appendix~\ref{app:ablations} the single-seed learning-rate spot check. Which simple strategy wins changes with the budget, but the family does not.

The affordable anchors are not close either. The six audited sophisticated selectors first become feasible at the median anchor, and across the sixteen anchors the mean winner exceeds the best affordable sophisticated method by $0.20$ to $7.12$\,pp. For Herding, the margin is not an accounting artifact: charging it zero selection cost, with every accuracy held fixed, leaves all sixteen winners unchanged (Appendix~\ref{app:ablations}). Its deficit is in accuracy at matched cost, not in the price of selection.

\subsection{ImageNet nearby-cost duels: full-data training is most accurate}
\label{sec:fixed-budget}

Compressed full-data training is the most accurate of the nearby-cost ImageNet points in both search groups, about 10 and 18 GPU-hours (Fig.~\ref{fig:duel}, Appendix~\ref{app:e2-full}), scored by final-epoch accuracy. In the higher group, full-data training reaches $66.17\%$, Uniform $65.83\%$, and Herding $64.12\%$ at nearly the same cost (Table~\ref{tab:e2-decomp}). The lower group orders the three the same way. Herding pays $5.45$h for selection before its first training epoch, roughly a third of the higher group's budget. At matched cost, both subset strategies trail the full-data point in both groups.

\subsection{Accuracy landscape: rankings are regime-dependent}
\label{sec:accuracy-landscape}

\begin{figure}[!ht]
  \centering
  \maybefig{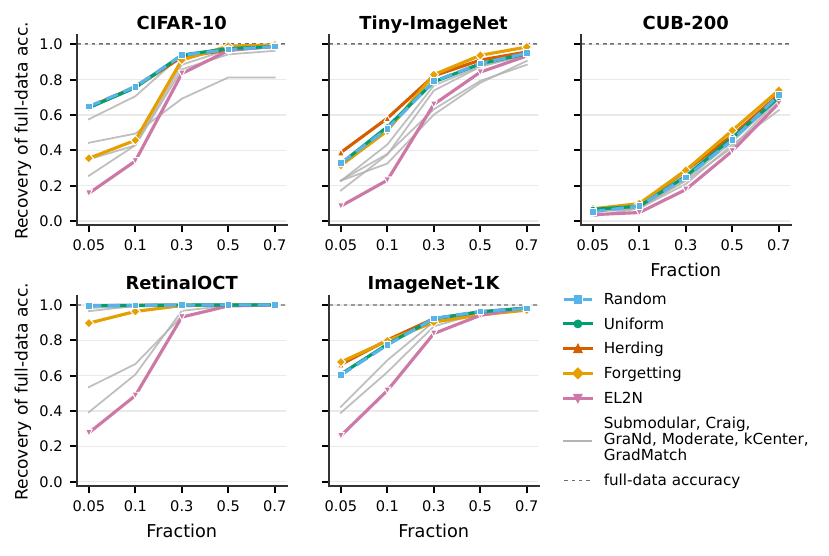}{0.78\linewidth}
  \caption{Recovery, the mean subset accuracy normalized by the mean full-data baseline, across methods, datasets, and fractions. Rankings depend on method and fraction.}
  \label{fig:recovery}
\end{figure}

Compared purely on downstream accuracy at fixed fractions, rankings depend on the fraction and no single method dominates, consistent with the pruning-regime dependence reported by \citet{sorscher2022beyond}. Fig.~\ref{fig:recovery} plots recovery, subset accuracy normalized by the full-data baseline, which compares accuracy loss across datasets of different difficulty.

At low fractions the ranking is decided by what a selector keeps: distribution matching leads and hardest-first ranking collapses. Herding leads Tiny ImageNet by more than $3$\,pp at both fractions $f{\le}0.1$ and tops ImageNet at $f{=}0.1$ (Fig.~\ref{fig:recovery}), which matches its distribution-matching motivation. EL2N, which in the main grid keeps the highest-scoring examples per class from epoch-1 scores, reaches only $14.98\%$ on CIFAR-10 at $f{=}0.05$, far below the worst of 30 Random subsets ($55.52\%$), and epoch-1 GraNd also underperforms. Scoring at epoch 20 instead narrows the gap without closing it, leaving EL2N $26$ to $30$\,pp below Random at both fractions (Appendix~\ref{app:score-epochs}). High-pruning failures of importance ranking have prior precedent \citep{sorscher2022beyond,zheng2023coverage}, but the magnitude reported here is specific to our scoring protocol. A feature-space check shows what the hardest-first rule keeps: EL2N and GraNd rank examples by early-training scores while respecting the class-balanced quota, and their selected ImageNet samples shift toward larger within-class distance percentiles (Appendix~\ref{app:feature-percentile}). kCenter shows a similar shift under a different rule.

At large fractions hard-example scoring often leads and the methods converge. Forgetting often ranks first at $f{\ge}0.5$, which matches hard-to-learn examples becoming more useful at larger fractions. Differences shrink: on the larger natural-image datasets most methods recover more than 90\% of full-data accuracy at $f{=}0.7$, CUB-200 excepted. Above half the data, the choice of selector matters little.

The same split transfers to a medical dataset where accuracy sits near its ceiling. On Retinal OCT, where the Full-data baseline is close to perfect, Moderate, Herding, Random, and Uniform stay within $1$\,pp of Full at $f{=}0.05$, while EL2N falls to $27.65\%$ and GraNd also collapses (Appendix~\ref{app:retinaloct}). Forgetting loses ground on this dataset, since Retinal OCT has many easy correctly classified images and few persistently forgotten ones. The Retinal OCT method ranking correlates with the natural-image average ranking at Spearman $\rho{=}0.66$, and the agreement is carried by the same split. The gap between coverage-preserving and hardest-first selection is not confined to natural images.

\subsection{End-to-end decomposition: selection cost does not extrapolate}
\label{sec:tradeoff}

Selection spends full-data computation before subset training begins, so the right counterfactual for a selector is the larger random subset the same time could train. As a reference, the 40 selector probe epochs alone are $20\%$ of the canonical 200-epoch full-data training count, before feature extraction and selection. The audit prices them per dataset, and Random's training-time curve converts each selection cost into the extra share of the training set that Random could train for it.

Priced this way, the largest fixed-fraction gain in the audited grid is bought more cheaply with more random data. On Tiny ImageNet at $f{=}0.1$, Herding gains $+3.91$\,pp over Random at $2.65\times$ Random's end-to-end time. Random would match that gain with about 4 more points of the training set, from $f{=}0.10$ to about $f{=}0.14$, while Herding's selection time pays for 20 more points, up to $f{=}0.30$. Choosing examples buys less accuracy than training on more of them.

The same arithmetic holds where the gain is small and the overhead smallest. On ImageNet at $f{=}0.3$, Herding gains essentially nothing over Random while its selection time would train Random on 16 more points. On CUB-200 at $f{=}0.3$, Herding gains $+1.97$\,pp at only $1.31\times$ Random's time, yet matching that gain needs roughly 4 more points while its selection time funds 32. In all three one-shot cases the same time buys more accuracy as a larger Random subset than as a selected one.

\begin{figure}[!ht]
  \centering
  \maybefig{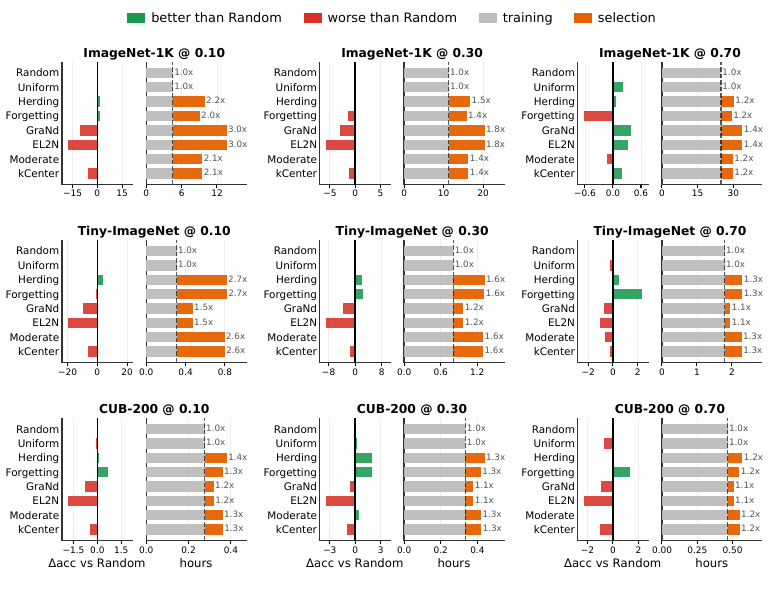}{0.80\linewidth}
  \caption{Fixed-fraction accuracy gain over Random beside the end-to-end
  time that gain costs, for ImageNet-1K, Tiny ImageNet and CUB-200 at
  $f\in\{0.1,0.3,0.7\}$. Left panels plot seed-mean accuracy minus Random's at
  the same fraction in pp, green above Random and red below. Right panels
  plot end-to-end hours split into training (grey) and selection (orange),
  with a dashed line at Random's end-to-end time, its multiple printed
  beside each bar, and per-cell axis scales.}
  \label{fig:gain-overhead}
\end{figure}

The pattern is general: Figure~\ref{fig:gain-overhead} puts gain beside price for every audited selector and fraction. Of the 54 priced scoring-selector cells, all cost more than Random at the same fraction, 34 also score below Random, and only 9 gain more than $1$\,pp. The epoch-1 scorers pay the most for the least: EL2N gives up $19.4$\,pp on Tiny ImageNet and $17.5$\,pp on ImageNet-1K at $f{=}0.1$ for $1.53\times$ and $3.03\times$ Random's time. A selector that costs more than Random and scores lower has no operating point worth choosing.

Selection cost is nearly fixed in the target fraction, because scores and embeddings are computed globally and only then sorted. At the smallest fraction it is therefore the larger part of end-to-end time: on ImageNet at $f{=}0.05$, selection consumes $62$ to $77\%$ of end-to-end time for the non-trivial audited selectors, and on Tiny ImageNet it spans $47$ to $73\%$. The smaller the subset, the larger the share of the budget that goes to choosing it.

\paragraph{The selection floor dominates at every scale.}
The floor dominates at every scale we measure: four-point fits of Herding's audited selection time against fraction put $95.3$ to $99.8\%$ of selection cost in the fraction-independent component on the four datasets (Appendix~\ref{app:timing-probe}). A smaller fraction therefore reduces downstream training time but leaves selection cost essentially untouched. What does not transfer across datasets is the floor's absolute size, about $234$\,s on CIFAR-10 and about $19{,}518$\,s on ImageNet-1K. Pricing selection in full-data epochs does not port either: per probe epoch, the $40$-epoch scorers cost about one full-data epoch on Tiny ImageNet but up to half again as much on CUB-200 and ImageNet-1K. The single-epoch scorers cost an order of magnitude more, because their charge includes ten independently trained scoring models. A multiplier fitted on CIFAR-scale measurements does not carry to ImageNet scale.

\paragraph{Robustness across architecture and training budget.}
The fixed-fraction ordering survives a change of backbone and a fourfold cut in training length. Two Tiny ImageNet ablations (all 11 methods, fractions $\{0.1, 0.3, 0.5\}$, 3 seeds) swap ResNet-18 for ViT-Tiny or cut training from 200 to 50 epochs. Rankings correlate with the main protocol at Spearman $\rho{=}0.94$ for short training and $\rho{=}0.78$ for ViT-Tiny (both $p{<}0.01$), and EL2N stays last at low fractions in all three settings (Table~\ref{tab:ablation-ranking}, Appendix~\ref{app:robustness}). Herding and Forgetting stay in the top tier under both ablations, while Craig and GradMatch move up under ViT-Tiny and kCenter drops to near-last. The top tier and the last place are stable, and only the middle ranks move.

\begin{table}[!ht]
  \caption{Tiny ImageNet average ranks (lower is better) under the standard ResNet-18 200-epoch protocol and the ResNet-18 50-epoch and ViT-Tiny 300-epoch ablations. Brackets give $95\%$ bootstrap intervals over seed resamples. Bold marks the rank-1 method in each column.}
  \label{tab:ablation-ranking}
  \centering
  \small
  \begin{tabular}{lccc}
    \toprule
    Method & ResNet-18, 200ep & ResNet-18, 50ep & ViT-Tiny, 300ep \\
    \midrule
    Random       & $3.56$ {\small [3.33,4.00]}   & $4.33$ {\small [3.78,5.00]}    & $4.61$ {\small [3.94,5.28]} \\
    Uniform      & $3.44$ {\small [3.11,3.78]}   & $4.67$ {\small [3.89,5.33]}    & $5.50$ {\small [4.89,6.17]} \\
    Herding      & $\mathbf{1.56}$ {\small [1.33,1.67]}   & $\mathbf{1.50}$ {\small [1.33,1.67]}    & $2.00$ {\small [2.00,2.00]} \\
    Submodular   & $5.00$ {\small [4.56,5.44]}   & $3.89$ {\small [3.11,4.44]}    & $4.33$ {\small [3.78,4.89]} \\
    Craig        & $10.56$ {\small [10.33,10.67]}& $10.67$ {\small [10.67,10.67]} & $6.94$ {\small [6.72,7.11]} \\
    Forgetting   & $2.56$ {\small [2.33,2.78]}   & $1.83$ {\small [1.50,2.17]}    & $\mathbf{1.00}$ {\small [1.00,1.00]} \\
    GraNd        & $8.11$ {\small [7.89,8.33]}   & $8.22$ {\small [8.00,8.33]}    & $8.44$ {\small [8.33,8.67]} \\
    EL2N         & $9.67$ {\small [9.67,9.67]}   & $9.67$ {\small [9.67,9.67]}    & $10.89$ {\small [10.67,11.00]} \\
    Moderate     & $4.89$ {\small [4.00,5.67]}   & $4.78$ {\small [4.11,5.44]}    & $5.00$ {\small [4.67,5.22]} \\
    kCenter      & $7.11$ {\small [7.00,7.33]}   & $7.11$ {\small [7.00,7.33]}    & $10.06$ {\small [9.89,10.28]} \\
    GradMatch    & $9.56$ {\small [9.33,9.78]}   & $9.33$ {\small [9.33,9.33]}    & $7.22$ {\small [7.00,7.33]} \\
    \bottomrule
  \end{tabular}
\end{table}

\subsection{Cost amortization with a fixed architecture}
\label{sec:amortization}

\begin{figure}[!ht]
  \centering
  \begin{subfigure}[t]{0.48\linewidth}
    \centering
    \maybefig{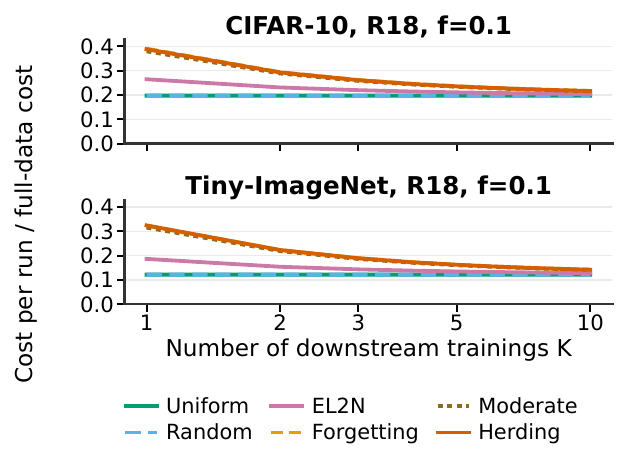}{\linewidth}
    \caption{Reuse amortization.}
    \label{fig:breakeven}
  \end{subfigure}\hfill
  \begin{subfigure}[t]{0.48\linewidth}
    \centering
    \maybefig{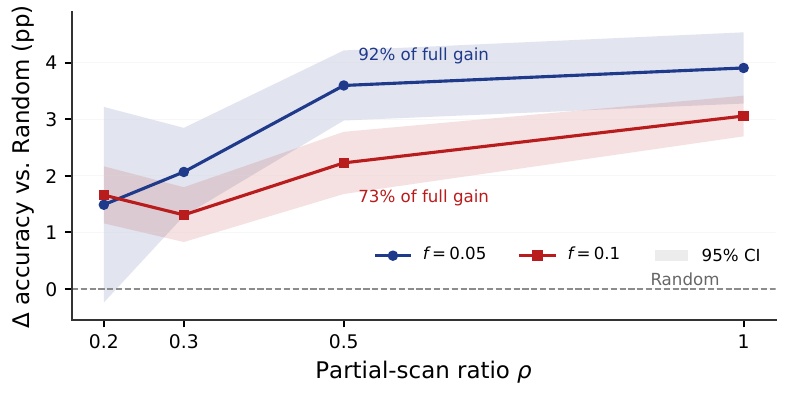}{\linewidth}
    \caption{Partial-scan Herding.}
    \label{fig:partial-scan-fig}
  \end{subfigure}
  \caption{Reuse and partial scanning. (a) R18-only cost per run against the
  number of reuses $K$ at $f{=}0.1$ under a fixed 200-epoch recipe, where the
  three 40-epoch scorers pay within $6\%$ of the same selection cost and so
  overlap. (b) Partial-scan Herding on Tiny ImageNet, 10 seeds per cell with
  $95\%$ CIs, plotting the gain over Random against the scanned fraction
  $\rho$ at the two smallest fractions.}
  \label{fig:reuse-row}
\end{figure}

Reuse makes a selected subset cheaper than full-data training after a single downstream run, but it does not close the accuracy gap. To isolate the cost effect, we fix selector and downstream architecture to ResNet-18 with the 200-epoch recipe on CIFAR-10 and Tiny ImageNet. If a subset is selected once and reused for $K$ trainings, its total cost is $c_{\rm sel}+K T(f)$, compared with $K T(1)$ for full-data training at the same epoch count, so the continuous cost crossover is $K^\ast=c_{\rm sel}/(T(1)-T(f))$ whenever $T(1)>T(f)$. All terms come from the R18 probes, and Fig.~\ref{fig:breakeven} plots the resulting cost per run. For the four learned selectors in Table~\ref{tab:r18-amortization} at $f\in\{0.1,0.3\}$, $K^\ast$ ranges from 0.07 to 0.31, and since actual $K$ is an integer they already cost less than a full 200-epoch run at $K=1$. Their subset accuracies stay below the full-data reference, so the crossover is in cost only.

Reuse does not, however, make a sophisticated selector cheaper than random selection. Random and Uniform obtain the same subset-training savings with negligible selection overhead, so under the common training-time model at fixed $f$ a sophisticated selector's extra selection cost per run shrinks as $1/K$ but never becomes a time saving relative to those baselines. Appendix~\ref{app:e5-table} reports R18-to-R50 accuracy transfer. Whatever value a selector has under reuse must come from accuracy.

\section{Discussion and Conclusion}
\label{sec:discussion}
\label{sec:conclusion}

\paragraph{Practical recommendations.}
At tight budgets, class-balanced random subsets at the largest affordable fraction are the strongest choice, compressed full-data schedules win several moderate and large anchors, and repeated random sampling is competitive where the pipeline allows it. Epoch-1 GraNd and EL2N perform poorly at low fractions, where the mean drop below Random reaches $46.9$\,pp (EL2N on CIFAR-10 at $f{=}0.05$). Under reuse, any accuracy premium should be compared against a same-size random subset, because amortization lowers cost per run but not accuracy.

\paragraph{Implications for future method design.}
For the audited selectors the full-dataset scan is the binding constraint, so the most direct improvement is to remove the scan rather than sharpen the score. A selector that does not beat a compressed full-data schedule at matched wall-clock cost has no one-shot efficiency advantage. A class-stratified partial-scan Herding pilot on Tiny ImageNet tests that route: feature extraction and per-class greedy matching use a random $\rho$ fraction of each class. Scanning half the data keeps $92\%$ of Herding's edge over Random at $f{=}0.05$ and $73\%$ at $f{=}0.10$ (Fig.~\ref{fig:partial-scan-fig}, Appendix~\ref{app:partial-scan}), and within-pilot mean selection time at $f{=}0.1$ falls from $3{,}876$\,s at $\rho{=}1$ to $841$\,s at $\rho{=}0.2$. Most of the edge survives a half scan, and selection time falls with $\rho$.

\paragraph{Limitations.}
The benchmark covers one-shot select-then-train pipelines with ResNet-18 recipes on four image datasets. All selection signals come from from-scratch training on the benchmark split, and every method uses its published selection hyperparameters under one shared recipe. Costs are single-host timing-probe estimates.

\paragraph{Conclusion.}
When each method is granted its best tested operating point, no budget anchor on either grid goes to a sophisticated selector, and reduced-epoch full-data training is the most accurate of the nearby-cost ImageNet points. Selection cost is dominated by a fixed full-dataset scan whose absolute size has to be measured again on every dataset, and a 2009 method that still tops our corrected fixed-fraction rankings wins no budget anchor once that cost enters the comparison. Selection time is part of the metric.

\subsubsection*{Ethics statement}
All experiments use public image classification datasets under their original licenses. No human subjects or personal data are involved. The correctness fixes we document concern a public codebase; we describe them as reproducibility issues, credit the original implementation, and do not question the integrity of prior results obtained with it.

\subsubsection*{Reproducibility statement}
All results derive from the released artifact bundle: corrected selector implementations, dataset preparation scripts, per-run result files with fixed seeds, selected-index files, raw timing-audit tables, and the analysis scripts that produce every figure and table in this paper. Section~\ref{sec:design} and Appendices~\ref{app:artifacts}, \ref{app:hparams-pointer}, and~\ref{app:timing-probe} document the protocol, hyperparameters, and audit methodology.

\subsubsection*{AI use statement}
In this work, we used generative AI tools to implement standard methods, to
assist with data analysis, and to draft and edit parts of the paper. We have reviewed all AI-assisted work: the
generated code was checked against the original DeepCore implementations and
tested by the authors, and all drafted and edited text was read and approved by the
authors. We take responsibility for the final content of this work, including
text, claims or artifacts produced with the aid of generative AI.

\bibliography{references}
\bibliographystyle{iclr2027_conference}

\appendix
\newpage

\section{Released Artifacts and Toolkit Contributions}
\label{app:artifacts}

This appendix documents the artifacts we release and the concrete increment our benchmark codebase provides over the original DeepCore \citep{guo2022deepcore} implementation.

\subsection{Released artifacts}
\label{app:artifacts-released}

The bundle carries the selected-index files, the raw numeric result CSVs, and the code. For each completed selection run, \texttt{indices.npy} holds the exact selected sample indices relative to the canonical training split, so follow-up work can evaluate robustness, distribution shift, or new training recipes directly on our selected subsets without re-running the selectors. One CSV per experiment family covers the main grid, the ViT and short-training ablations, the budget-grid and fixed-budget-duel runs, the reuse study, and the partial-scan pilot, plus metadata manifests, and each CSV holds accuracy, selection time, training time, end-to-end time, best epoch, and selected count. The raw CSVs label kCenter with its historical method ID \texttt{CCS}, so file names and hashes stay stable across releases. The code ships corrected selector implementations, dataset loaders, experiment schedulers, timing-audit scripts, analysis and plotting scripts, fixed random seeds, dataset preparation instructions, and the environment specification. Raw image datasets are not redistributed.

\subsection{Bug fixes in inherited code}
\label{app:bugfixes}

Table~\ref{tab:bugfixes} lists the 9 correctness issues we fixed, plus one protocol change. The most consequential is a sign error in Herding's greedy selection loop: the original code selects the sample \emph{farthest} from the running surrogate mean (\texttt{argmax}), whereas Herding's construction requires the \emph{nearest} sample (\texttt{argmin}). Fixing this bug alone improves CIFAR-10 Herding accuracy at $f{=}0.1$ from $66.36\%$ to $72.24\%$, a $+5.9$\,pp shift over 3 seeds under the unified recipe. Herding is a common baseline, so rankings computed with the original code are off by $5.9$\,pp on this cell. Two further fixes change accuracy: the GraphCut marginal-gain term affects Submodular, and the dropped per-sample weights affect the weighted selectors Craig and GradMatch, whose subsets the original loader trained unweighted. Every released run uses the corrected code. The remaining fixes are feasibility fixes: without them the code does not run on current PyTorch and rejects $64\times64$ inputs.

\begin{table}[h]
  \caption{Fixes applied to the inherited DeepCore code. Impact marks accuracy-changing versus feasibility issues.}
  \label{tab:bugfixes}
  \centering
  \footnotesize
  \begin{tabular}{@{}p{0.27\linewidth}p{0.43\linewidth}p{0.23\linewidth}@{}}
    \toprule
    Location & Bug & Impact \\
    \midrule
    \texttt{herding.py:82} & Greedy selects farthest point (\texttt{argmax}) instead of nearest (\texttt{argmin}) & Accuracy (${\sim}{+}5.9$\,pp on CIFAR-10) \\
    \texttt{submodular\_fn.py:105} & GraphCut $\lambda$ on wrong term of marginal gain & Accuracy \\
    \texttt{craig.py:48} & Dereferences non-existent \texttt{dst\_val.targets} attribute & Feasibility (crash) \\
    \texttt{gradmatch.py:68} & Calls \texttt{torch.lstsq}, removed in PyTorch 2.x & Feasibility (crash) \\
    \texttt{gradmatch.py} (OMP loop) & \texttt{torch.cat} in loop fragments CUDA memory & Feasibility (OOM on large classes) \\
    \texttt{resnet.py:105} & Hard-coded \texttt{avg\_pool2d(out, 4)} breaks 64$\times$64 inputs & Feasibility (shape error on Tiny ImageNet) \\
    \texttt{utils.py} (\texttt{WeightedSubset}) & Inherits \texttt{Subset}, DataLoader drops per-sample weights & Accuracy (weighted methods silently unweighted) \\
    \texttt{utils.py} (\texttt{train}) & Nested indexing in weighted branch produces 3D inputs & Feasibility (crash in weighted training) \\
    \texttt{forgetting.py:36} & Per-batch index tensor on CPU, score tensor on GPU & Feasibility (device-mismatch crash) \\
    \texttt{run\_experiments.py} (GraNd, EL2N) & Default \texttt{selection\_epochs=40} for rules introduced as ``early-training'' scores & Protocol change, not a defect \\
    \bottomrule
  \end{tabular}
\end{table}

\subsection{Newly added methods, datasets, and infrastructure}
\label{app:new-methods}

We added EL2N \citep{paul2021deep}, Moderate \citep{xia2023moderate}, and class-balanced greedy kCenter as first-class selectors under the same \texttt{CoresetMethod} interface, re-implemented Random with global unbalanced semantics strictly distinct from class-balanced Uniform, and integrated CUB-200 and Retinal OCT with canonical loaders and dataset-specific augmentation routing. New infrastructure includes the multi-GPU grid scheduler with resume semantics, the ablation scheduler, the single-machine core-pinned timing audit with a per-minute resource monitor, ImageNet preflight and launch tooling, the budget-grid and duel launchers, and the analysis pipeline that produces every figure and table. The inherited entry point supported only single-shot training with no end-to-end accounting.

\subsection{Pointer to the full hyperparameter tables}
\label{app:hparams-pointer}

Table~\ref{tab:hparams} lists the downstream training protocol per dataset. All datasets share SGD with momentum $0.9$, Nesterov acceleration, weight decay $5\times10^{-4}$, and cosine annealing to the listed final learning rate. Train-time augmentation is one fixed policy per dataset: a random crop with reflection padding of 4 pixels on CIFAR-10 and 8 pixels on Tiny ImageNet followed by a horizontal flip, and a random resized crop to $224$ followed by a horizontal flip on CUB-200, ImageNet-1K, and Retinal OCT. Evaluation applies no augmentation: CIFAR-10 and Tiny ImageNet are normalized at native resolution, and CUB-200, ImageNet-1K, and Retinal OCT use a deterministic resize to $256$ with a center crop to $224$. The policy is fixed across methods and fractions, so no method is advantaged by its own recipe. The ViT ablation uses a DeiT-style recipe documented in Appendix~\ref{app:robustness}. Per-method selection hyperparameters follow the corrected DeepCore defaults and are maintained as machine-readable dictionaries in the released schedulers.

\begin{table}[h]
  \caption{Downstream training protocol per dataset (fixed across all methods and fractions).}
  \label{tab:hparams}
  \centering
  \footnotesize
  \begin{tabular}{lccccc}
    \toprule
    Dataset & Batch & Peak lr & Final lr & Epochs & Train-time input pipeline \\
    \midrule
    CIFAR-10      & 256 & $0.1$  & $10^{-4}$ & 200 & random crop 32 (pad 4), flip, normalize \\
    Tiny ImageNet & 128 & $0.1$  & $10^{-4}$ & 200 & random crop 64 (pad 8), flip, normalize \\
    CUB-200       & 64  & $0.01$ & $10^{-6}$ & 200 & random resized crop 224, flip, normalize \\
    ImageNet-1K   & 256 & $0.1$  & $10^{-4}$ & 90  & random resized crop 224, flip, normalize \\
    \bottomrule
  \end{tabular}
\end{table}

\section{Methods Outside the Main Grid}
\label{app:scope}

\emph{Selection via Proxy}~\citep{coleman2020selection} fits the fixed-subset protocol and bears on selection cost because it changes proxy architecture and training duration. The grid does not cover that design axis. The original Coverage-centric Coreset Selection algorithm \citep{zheng2023coverage} also fits the protocol, but requires an importance score, score strata, and a tuned hardest-sample cutoff. It is a different method from our kCenter baseline and is not in the grid. \emph{$\mathbb{D}^2$ Pruning}~\citep{maharana2024d2} likewise fits the decomposition but was outside the completed grid. \emph{GLISTER}~\citep{killamsetty2021glister} couples selection with the inner training loop, while \emph{DQ}~\citep{zhou2023data} trains on synthesized proxy data, so neither matches our static-subset primitive. \emph{MoSo}~\citep{tan2023moso} is intractable in exact form (the authors estimate $>45$ years on a V100 for ImageNet-1K). \emph{SAGE}~\citep{jha2025sage} is a streaming gradient-sketch selector that the grid does not include.

\section{Budget-Grid Details and Full Tables}
\label{app:e1-full}

\begin{figure}[!ht]
  \centering
  \maybefig{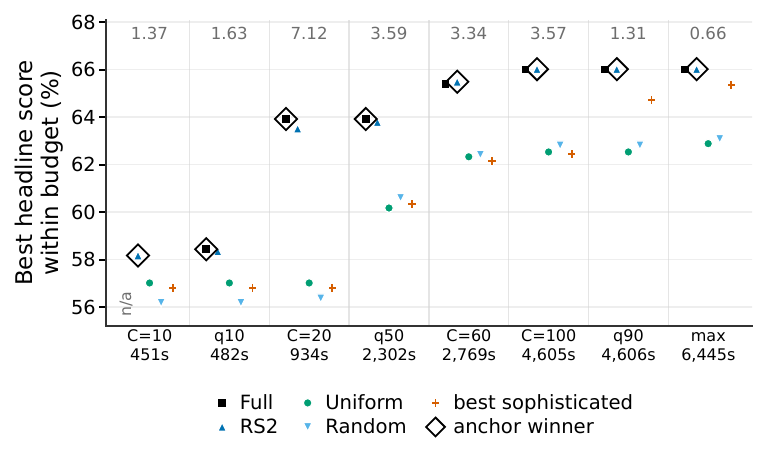}{0.95\linewidth}
  \caption{Tiny ImageNet budget anchors, the counterpart of
  Fig.~\ref{fig:budget-grid}. The audited selection floors are seven to
  eight times larger than on CIFAR-10 (about $10$ minutes for GraNd and EL2N
  and $29$ to $31$ minutes for the other audited selectors), so the audited
  sophisticated selectors are absent from the affordable set until the
  median anchor. Full-data
  training has no affordable operating point at the cheapest anchor, marked
  n/a.}
  \label{fig:budget-grid-tiny}
\end{figure}

Anchor budgets are computed on the released measured grid (357 CIFAR-10 runs over 121 cells, 118 with all 3 seeds, and 327 Tiny ImageNet runs over 109 cells, all with 3 seeds), with end-to-end cost per cell equal to mean training time plus that method's audited selection cost, which is flat in fraction. The strategy pool adds the canonical 200-epoch cells of every method (55 on CIFAR-10 and 56 on Tiny ImageNet after deduplication), so full-data training at its canonical schedule and every sophisticated selector's canonical cell compete at the anchors they can afford. Those cells enter the pool but not the anchor construction, which is why the maximum anchor on either dataset does not reach the canonical full-data cell. The full-data cells use a linear extrapolation of the controlled training probes, checked against a direct probe in Appendix~\ref{app:full-probe-check}. Winners use three-seed mean recorded-final accuracy. Missing-final sophisticated canonical cells retain peak as a favorable upper bound, while missing-final simple canonical cells are excluded. Tables~\ref{tab:e1-cifar} and~\ref{tab:e1-tiny} report each anchor's winner, operating point, runner-up, and margin. Every winner on either dataset is a simple strategy. The sophisticated runner-ups are Submodular at the two cheapest CIFAR-10 anchors ($84.88$) and GraNd at its median anchor ($90.80$), and all Tiny ImageNet runner-ups are simple. Submodular is charged zero selection cost, its most favorable pricing, and its runner-up cell has $0.009$\,s and $13.67$\,s of budget slack at these two anchors. The single-seed learning-rate cell excluded at the CIFAR-10 q10 anchor is itself a Uniform variant, so that anchor goes to class-balanced random sampling under either eligibility rule. Figure~\ref{fig:budget-grid-tiny} shows the Tiny ImageNet anchor chart.

The sensitivity checks below are computed on peak accuracy, on the same grid and the same cost basis as the headline analysis. Exact enumeration of the 27 ordered seed resamples produces sophisticated winners in 12 of 216 CIFAR-10 anchor decisions: Submodular wins 8 resamples at $C=10$ and 1 at $C=20$, and GraNd wins 3 at q50. Tiny ImageNet has no such reversal in 216 decisions. Submodular retains the favorable zero-selection-cost bound. The eight anchors per dataset share runs, and some anchors share a budget.

Common selection/training multipliers from $\{0.9,1,1.1\}^2$ change no peak-based winners in the 128 non-baseline anchor decisions. Independent per-cell perturbations change more decisions: 2,000 draws with each selection and training component multiplied by an independent $\mathrm{Unif}(0.9,1.1)$ factor yield at least one sophisticated winner in 361 draws on CIFAR-10 and 167 on Tiny ImageNet, recomputing anchors each time. An adversary that makes sophisticated cells $10\%$ cheaper and simple cells $10\%$ dearer also reverses decisions at fixed anchors. Some anchors sit exactly on a simple strategy's feasibility boundary, so even a small increase can exclude it. The main result is the unperturbed three-seed mean final/favorable-bound comparison. The same cost perturbation repeated with the headline final/favorable-bound scores yields at least one sophisticated winner in $332/2000$ independent perturbation draws on CIFAR-10 and $172/2000$ on Tiny ImageNet. No sophisticated winner appears under the nine common-factor settings on either dataset. The artifact script \texttt{check\_primary\_cost\_sensitivity.py} reproduces this check.

\begin{table}[!ht]
  \caption{CIFAR-10 anchors, priced on the single-machine timing audit: winner and runner-up by three-seed mean headline score among affordable cells. Scores use recorded final accuracy, with the favorable-bound treatment for missing sophisticated canonical finals. The single-seed lr 0.05 ablation reaches a peak of 87.77 at the q10 budget but is not eligible (Appendix~\ref{app:ablations}). Submodular is priced at zero selection cost.}
  \label{tab:e1-cifar}
  \centering
  \small
  \begin{tabular}{lcllclc}
    \toprule
    Anchor & Budget (s) & Winner & $(f, E)$ & Score (\%) & Runner-up & Margin (pp) \\
    \midrule
    $C{=}10$  & 64  & Uniform & $(0.7, 14)$  & 85.21 & Submodular & 0.33 \\
    q10       & 78  & Uniform & $(0.7, 14)$  & 85.21 & Submodular & 0.33 \\
    $C{=}20$  & 132 & Full    & $(1.0, 20)$  & 91.00 & RS2        & 0.57 \\
    q50       & 351 & Full    & $(1.0, 20)$  & 91.00 & GraNd      & 0.20 \\
    $C{=}60$  & 393 & Full    & $(1.0, 60)$  & 94.18 & RS2        & 0.47 \\
    $C{=}100$ & 653 & Full    & $(1.0, 100)$ & 94.82 & RS2        & 0.15 \\
    q90       & 653 & Full    & $(1.0, 100)$ & 94.82 & RS2        & 0.15 \\
    max       & 913 & RS2     & $(0.7, 200)$ & 95.01 & Full       & 0.19 \\
    \bottomrule
  \end{tabular}
\end{table}

\begin{table}[!ht]
  \caption{Tiny ImageNet anchors, priced on the single-machine timing audit, using the same final/favorable-bound score as Table~\ref{tab:e1-cifar}. All eight go to RS2 or full-data training, with the other of the two as runner-up at seven of eight. Uniform is the runner-up at the cheapest anchor.}
  \label{tab:e1-tiny}
  \centering
  \small
  \begin{tabular}{lcllclc}
    \toprule
    Anchor & Budget (s) & Winner & $(f, E)$ & Score (\%) & Runner-up & Margin (pp) \\
    \midrule
    $C{=}10$  & 451  & RS2  & $(0.7, 14)$  & 58.17 & Uniform & 1.16 \\
    q10       & 482  & Full & $(1.0, 10)$  & 58.44 & RS2     & 0.09 \\
    $C{=}20$  & 934  & Full & $(1.0, 20)$  & 63.92 & RS2     & 0.40 \\
    q50       & 2302 & Full & $(1.0, 20)$  & 63.92 & RS2     & 0.13 \\
    $C{=}60$  & 2769 & RS2  & $(0.7, 86)$  & 65.49 & Full    & 0.10 \\
    $C{=}100$ & 4605 & RS2  & $(0.7, 143)$ & 66.02 & Full    & 0.01 \\
    q90       & 4606 & RS2  & $(0.7, 143)$ & 66.02 & Full    & 0.01 \\
    max       & 6445 & RS2  & $(0.7, 143)$ & 66.02 & Full    & 0.01 \\
    \bottomrule
  \end{tabular}
\end{table}

\section{Accuracy Basis and Peak Comparison}
\label{app:final-acc}

Selecting a run's peak test accuracy chooses an epoch using the test set. Grid runs record both peak and final accuracy, and their mean difference is $0.10$\,pp on CIFAR-10 and $0.12$\,pp on Tiny ImageNet. The headline E1 analysis uses recorded final accuracy. Historical canonical finals are accepted only for unmodified runs whose logged peak matches the canonical result, and repaired runs do not inherit old finals. Missing sophisticated finals retain their peak as a favorable upper bound, and missing simple-family finals are excluded. All 684 E1 grid runs have finals. Among the additional canonical 200-epoch cells merged into the strategy pool, 14 CIFAR-10 cell means and 24 Tiny ImageNet cell means lack a verified final: CIFAR-10 has Craig 5, GradMatch 5, Random 3, and Full 1, and Tiny ImageNet has kCenter 5, Craig 5, EL2N 5, GradMatch 5, Moderate 3, and Full 1.

Table~\ref{tab:final-acc} contrasts these headline scores with the historical peak analysis at the same anchors. Relative to peak, no CIFAR-10 winner changes and two Tiny ImageNet winners change from Full to RS2, and no winner leaves the simple family. The smallest simple-over-sophisticated margin is $0.20$\,pp on CIFAR-10 and $0.66$\,pp on Tiny ImageNet. The shared analysis is implemented in \texttt{e1\_analysis.py}.

\begin{table}[!ht]
\caption{Historical peak scores versus the headline recorded-final/favorable-bound scores at the same audited anchors. Gap is the headline winner minus the best affordable sophisticated method under the headline rule.}
\label{tab:final-acc}
\centering\small
\begin{tabular}{lrrrrr}
\toprule
Anchor & Peak winner & Peak acc. & Headline winner & Score (\%) & Gap (pp) \\
\midrule
\multicolumn{6}{l}{\emph{CIFAR-10}} \\
C=10 & Uniform & 85.21 & Uniform & 85.21 & 0.33 \\
q10 & Uniform & 85.21 & Uniform & 85.21 & 0.33 \\
C=20 & Full & 91.06 & Full & 91.00 & 0.79 \\
q50 & Full & 91.06 & Full & 91.00 & 0.20 \\
C=60 & Full & 94.21 & Full & 94.18 & 0.71 \\
C=100 & Full & 94.88 & Full & 94.82 & 0.75 \\
q90 & Full & 94.88 & Full & 94.82 & 0.75 \\
max & RS2 & 95.09 & RS2 & 95.01 & 0.47 \\
\midrule
\multicolumn{6}{l}{\emph{Tiny ImageNet}} \\
C=10 & RS2 & 58.25 & RS2 & 58.17 & 1.37 \\
q10 & Full & 58.44 & Full & 58.44 & 1.63 \\
C=20 & Full & 63.98 & Full & 63.92 & 7.12 \\
q50 & Full & 63.98 & Full & 63.92 & 3.59 \\
C=60 & RS2 & 65.55 & RS2 & 65.49 & 3.34 \\
C=100 & Full & 66.12 & RS2 & 66.02 & 3.57 \\
q90 & Full & 66.12 & RS2 & 66.02 & 1.31 \\
max & RS2 & 66.27 & RS2 & 66.02 & 0.66 \\
\midrule
\end{tabular}\end{table}

\section{ImageNet Duel Operating Points}
\label{app:e2-full}

Table~\ref{tab:e2-decomp} prices the tested ImageNet operating points entirely from the controlled R18 timing probes. Selection uses the mean over audited fractions, $f{=}0.5$ training is interpolated, and $f{=}1$ training is extrapolated. The 10h and 18h labels identify the original search groups rather than exact caps. Moderate, kCenter, and EL2N appear only in the higher group, and GraNd was not probed. The accuracy comparison covers these operating points.

Full has the highest accuracy in both groups and the lowest estimated cost at the nominal probe values. Its cost lead over the next cheapest point is only $0.13\%$ in the higher group and $3.10\%$ in the lower group. Common selection/training perturbations of $\pm10\%$ preserve its cheapest status in 6 of 9 settings per group.

\begin{table}[!ht]
\caption{ImageNet operating points priced from the timing probes. Final-epoch accuracy is mean $\pm$ population standard deviation over three seeds, and daggers mark one seed. Uniform's nonzero indexing cost rounds to 0.00h.}
\label{tab:e2-decomp}
\centering\small
\begin{tabular}{llrrrrr}
\toprule
Group & Strategy & $(f,E)$ & Sel. (h) & Train (h) & Total (h) & Top-1 (\%) \\
\midrule
18h & Full & (1,44) & 0.00 & 17.11 & 17.11 & $66.17\pm0.07$ \\
18h & Uniform & (0.7,63) & 0.00 & 17.36 & 17.36 & $65.83\pm0.05$ \\
18h & Herding & (0.5,59) & 5.45 & 11.83 & 17.28 & $64.12\pm0.06$ \\
18h & Moderate & (0.5,62) & 5.01 & 12.43 & 17.44 & $63.98^\dagger$ \\
18h & kCenter & (0.5,62) & 5.01 & 12.43 & 17.44 & $63.87^\dagger$ \\
18h & Forgetting & (0.5,64) & 4.56 & 12.83 & 17.39 & $63.24^\dagger$ \\
18h & EL2N & (0.5,40) & 9.12 & 8.02 & 17.14 & $61.30^\dagger$ \\
10h & Full & (1,26) & 0.00 & 10.11 & 10.11 & $64.77\pm0.13$ \\
10h & Uniform & (0.5,52) & 0.00 & 10.43 & 10.43 & $63.76\pm0.01$ \\
10h & Herding & (0.3,40) & 5.45 & 5.02 & 10.46 & $59.94^\dagger$ \\
10h & Forgetting & (0.3,47) & 4.56 & 5.89 & 10.46 & $59.54^\dagger$ \\
\bottomrule\end{tabular}\end{table}

\section{Same-Architecture Amortization and Separate Transfer Evidence}
\label{app:e5-table}

Table~\ref{tab:r18-amortization} holds the architecture and epoch budget fixed. It uses controlled R18 selection costs and R18 training probes at $f{=}0.1,0.3$, with full-data training linearly extrapolated to $f{=}1$. The continuous crossover is below one for these cells, so the first whole training already costs less than a full 200-epoch run.

\begin{table}[!ht]
\caption{R18 cost crossover $K^\ast$ versus full-data R18 training at 200 epochs. Actual reuse counts are integers.}
\label{tab:r18-amortization}
\centering\small
\begin{tabular}{lrrrr}
\toprule
& \multicolumn{2}{c}{CIFAR-10} & \multicolumn{2}{c}{Tiny ImageNet} \\
Method & $f{=}0.1$ & $f{=}0.3$ & $f{=}0.1$ & $f{=}0.3$ \\
\midrule
EL2N & 0.08 & 0.11 & 0.07 & 0.10 \\
Forgetting & 0.24 & 0.31 & 0.23 & 0.30 \\
Moderate & 0.23 & 0.29 & 0.22 & 0.28 \\
Herding & 0.24 & 0.31 & 0.23 & 0.30 \\
\bottomrule\end{tabular}\end{table}

The R18-to-R50 reuse experiment instead tests architecture transfer (Table~\ref{tab:e5-reuse}). It reuses the selected indices to train R50 and reports accuracy relative to R50 trained on same-size random subsets. The table reports accuracy only, since the audit has no R50 probes. CIFAR-10 rows use three seeds and Tiny ImageNet rows two. Herding retains an advantage on Tiny ImageNet, while the premium depends on dataset, fraction, and comparator.

\begin{table}[!ht]
\caption{Separate R18-to-R50 transfer experiment: accuracy difference (pp) from R50 trained on same-size random subsets.}
\label{tab:e5-reuse}
\centering\small
\begin{tabular}{lrrrr}
\toprule
& \multicolumn{2}{c}{CIFAR-10} & \multicolumn{2}{c}{Tiny ImageNet} \\
Method & $f{=}0.1$ & $f{=}0.3$ & $f{=}0.1$ & $f{=}0.3$ \\
\midrule
Uniform & $+2.2$ & $+1.5$ & $+1.8$ & $+0.0$ \\
Herding & $+2.0$ & $+1.5$ & $+5.1$ & $+2.5$ \\
Moderate & $-0.8$ & $+0.7$ & $+2.5$ & $-0.7$ \\
Forgetting & $-26.1$ & $-0.0$ & $+1.1$ & $+3.1$ \\
EL2N & $-46.4$ & $-6.8$ & $-19.1$ & $-10.8$ \\
\bottomrule\end{tabular}\end{table}

\section{Feature-Space Check for Low-Fraction Failures}
\label{app:feature-percentile}

\begin{figure}[h]
  \centering
  \maybefig{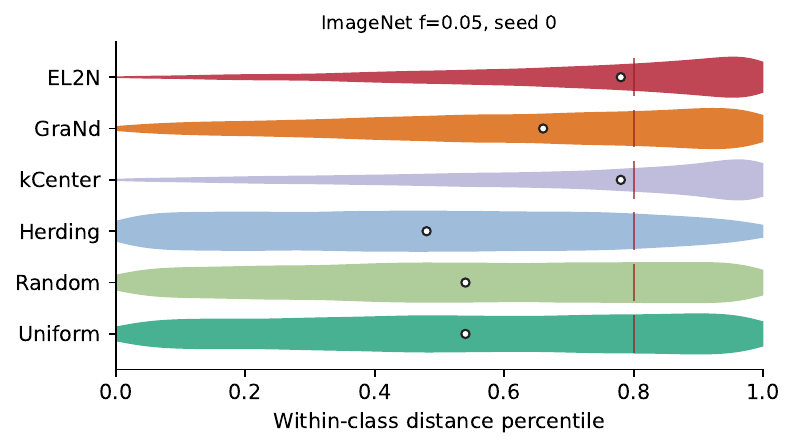}{0.55\linewidth}
  \caption{ImageNet $f{=}0.05$ feature-distance percentiles. Width indicates selected-sample density after converting each image's distance to its class centroid into a within-class percentile. Dots mark medians and red ticks mark the 80th percentile. EL2N, GraNd, and kCenter shift toward atypical samples.}
  \label{fig:feature-percentile}
\end{figure}

We embed the full ImageNet-1K training split with an ImageNet-pretrained ResNet-50 checkpoint, classification head removed, giving 2048-dimensional penultimate features. The extractor is external to the benchmark, since selector-probe features would make the diagnostic circular. It is used for analysis only, not for selection. For each image we compute the Euclidean distance to its class centroid and convert it into a within-class percentile against the class's full distance distribution, so a method that samples a class uniformly produces a flat percentile profile. Figure~\ref{fig:feature-percentile} shows these percentiles for a 10{,}000-image subsample of each method's fraction-0.05 selection (seed 0). Random and Uniform sit near the flat reference (median percentile 0.54), and Herding shifts modestly toward class cores (median 0.48). GraNd's median percentile is 0.66 and kCenter's and EL2N's are 0.78, with $48\%$ of kCenter and EL2N selections beyond the 80th percentile against the $20\%$ a flat profile would give. All three selectors under-sample class cores at this fraction.

\section{Ablations: Selection Probe, Schedule Cap, and Learning Rate}
\label{app:ablations}

Five checks examine the evaluated recipe: a cheaper selection probe, a hypothetical cheaper Herding charge, later scoring epochs for EL2N, uncapped schedules, and the learning rate.

\paragraph{Cheaper selection probes (se10).}
Running Tiny ImageNet Herding with a 10-epoch rather than 40-epoch selector probe leaves accuracy unchanged or slightly better ($39.54 \pm 0.33$ against $38.73 \pm 0.31$ at $f{=}0.1$, $54.66 \pm 0.09$ against $54.72 \pm 0.42$ at $f{=}0.3$, 3 seeds each). These runs carry accuracy only, since their clocks are ordinary run clocks rather than audit probes.

\paragraph{Hypothetical cheaper Herding.}
To separate accounting sensitivity from that missing measurement, we multiply Herding's entire audited selection charge by $\alpha\in\{1,0.25,0\}$, retaining all existing headline accuracies, training costs, other selectors' costs, and the original 16 budget anchors. All 16 winners remain unchanged, even at zero Herding selection cost. Recomputing the anchors gives the same winner result. On Tiny ImageNet, cheaper Herding becomes the best sophisticated alternative at some anchors, but never a winner. The quarter-cost scenario scales extraction and greedy work as well, so it is an upper bound on what a cheaper probe could save. The script \texttt{analyze\_b\_herding\_cost\_scenarios.py} exports all decisions and boundary slacks.

\paragraph{Later scoring epochs for EL2N.}
\label{app:score-epochs}
The main grid scores GraNd and EL2N at epoch 1, earlier than the epoch-10 and epoch-20 settings used by \citet{paul2021deep}. To check whether the low-fraction collapse is confined to that early scoring point, we rerun CIFAR-10 EL2N with scoring epochs $t\in\{1,10,20\}$ at $f\in\{0.05,0.1\}$ over seeds $\{0,1,2\}$, together with Random and Uniform controls in the same batch. Each selection averages scores from ten independently initialized ResNet-18 probes and keeps the highest-scoring examples within each class. The downstream recipe is unchanged (fresh ResNet-18, 200 epochs). Table~\ref{tab:score-epochs} reports final accuracy. In the main grid, epoch-1 EL2N reaches $14.98\%$ at $f{=}0.05$, $40.5$\,pp below the worst of 30 Random subsets ($55.52\%$). Later scoring raises accuracy monotonically, to $32.13\%$ and $46.44\%$ at epoch 20 for $f{=}0.05$ and $0.1$, and the epoch-20 subsets remain $29.8$ and $26.3$\,pp below Random at the two fractions. These reruns carry accuracy only.

\begin{table}[ht]
  \caption{CIFAR-10 EL2N with later scoring epochs: final test accuracy, mean $\pm$ sample SD over three seeds, controls rerun in the same batch.}
  \label{tab:score-epochs}
  \centering
  \footnotesize
  \begin{tabular}{@{}lcc@{}}
    \toprule
    Selection & $f{=}0.05$ & $f{=}0.1$ \\
    \midrule
    EL2N, scored at epoch 1 & $14.12 \pm 1.46$ & $28.80 \pm 5.95$ \\
    EL2N, scored at epoch 10 & $25.07 \pm 0.13$ & $38.23 \pm 1.72$ \\
    EL2N, scored at epoch 20 & $32.13 \pm 1.35$ & $46.44 \pm 1.55$ \\
    Random & $61.96 \pm 0.90$ & $72.76 \pm 0.77$ \\
    Uniform & $60.72 \pm 2.71$ & $72.66 \pm 1.50$ \\
    \bottomrule
  \end{tabular}
\end{table}

\paragraph{Uncapped schedules.}
The grids cap epochs at 200, which could in principle deny small-fraction selection its best regime. Uncapped probes at (0.05, 400), (0.1, 300), and (0.1, 400) on CIFAR-10 do not support that: the best of them (Herding at $f{=}0.1$ for 400 epochs, $74.44 \pm 1.15$) remains more than $16$\,pp below what full-data training reaches at a comparable end-to-end cost, and Herding's uncapped cells do not separate from Uniform's ($74.63 \pm 0.62$ at the same operating point). Long schedules on small selected subsets saturate far below the anchors they would need to contest.

\paragraph{Learning rate.}
All grid cells share the canonical lr 0.1. A spot check at Uniform (0.5, 20) finds lr 0.05 reaching $87.77$ against $81.80 \pm 0.72$ at lr 0.1 and $81.07$ at lr 0.2 (single seed for the variants). Short-schedule operating points are undertuned for every strategy, and tuning helps the simple strategy that already wins the low anchors. Because the lr cells are single-seed and were not granted to other methods, they are not eligible to win in Table~\ref{tab:e1-cifar}. The lr 0.05 cell would otherwise take the q10 anchor by $2.56$\,pp over the standard-recipe winner.

\section{Timing-Audit Probe Details}
\label{app:timing-probe}

All training-time estimates are produced by a per-dataset probe recorded in the released audit table. Each row contains the tuple (dataset, fraction, warmup epochs, timed epochs, test repeats, target epochs, eval calls, mean train-epoch time, mean test-pass time, estimated training time). The headline protocol uses 2 warmup epochs, 10 timed epochs, and 3 test repeats for every reported pair, with the estimator
\[
\text{estimated\_training\_time} \;=\; \text{target\_epochs} \cdot \overline{t_{\text{train}}} \;+\; \text{target\_epochs} \cdot q \cdot \overline{t_{\text{test}}}.
\]
The audit uses $q=110/200$ evaluations per epoch on the small datasets and $q=49/90$ on ImageNet. This rate is retained for shortened schedules to keep one cost convention, and it approximates their integer evaluation counts. Selection costs are means over the audited fractions. Four-point fits of Herding's audited selection time against fraction give ${\approx}234 + 16 f$ seconds on CIFAR-10, ${\approx}378 + 2 f$ on CUB-200, ${\approx}1{,}839 + 4 f$ on Tiny ImageNet, and ${\approx}19{,}518 + 339 f$ on ImageNet-1K, with the fraction-independent component accounting for $95.3$, $99.6$, $99.8$, and $98.8\%$ of selection cost respectively. The ratio of audited selection cost to probe epochs times one audited full-data epoch is $0.99$ to $1.03$ on Tiny ImageNet against $1.21$ to $1.52$ on CUB-200 and ImageNet-1K for the $40$-epoch scorers, and $13$ to $25$ for the single-epoch scorers. Exact training fractions use their measured probe rates, $f{=}0.5$ uses a least-squares interpolation, and $f{=}1$ uses the same linear fit extrapolated beyond the measured range. In-range fit quality does not certify the direction or size of the full-data extrapolation error. The CIFAR-10 extension used the same host and affinity settings. The sensitivity analyses describe dependence on the estimates, not a calibrated measurement-error distribution.

The 10-epoch timed window captures within-run jitter from data-loader warmup, caching, and I/O contention, which biases short probes more on ImageNet-scale data. End-to-end time is selection time (measured directly, wall-clock, on an isolated GPU) plus this probe-based training estimate. We release the raw probe CSVs and protocol metadata so readers can reconstruct the estimator.

\subsection{Direct full-data check}
\label{app:full-probe-check}

A separate follow-up measures full-data training on the same host, using the same model, batch sizes, augmentation, worker count, thread limits, 8-logical-CPU affinity blocks, and 12-epoch probe learning-rate schedule. It uses 2 warmup epochs and 3 test passes, with a timed window of 10 epochs on CIFAR-10, Tiny ImageNet, and CUB-200 and 3 epochs on ImageNet. The affinity blocks and GPU assignments need not match the historical assignments. Full-data mean training-epoch times are 5.965, 44.784, 6.276, and 1374.202 seconds respectively, differing from the original linear extrapolations by $+1.79\%$, $+0.40\%$, $+1.13\%$, and $+0.86\%$. Repeated $f{=}0.7$ controls on the three smaller datasets differ from their historical measurements by $+2.47\%$, $+0.26\%$, and $+0.72\%$. These are observed discrepancies, not confidence bounds on measurement error.

Substituting the new full-data training and evaluation rates leaves all sixteen E1 winners in the simple-strategy family, both with the original anchor budgets and after recomputing the anchors. On ImageNet, the Full points move from 10.11/17.11h to 10.20/17.26h, their accuracies are unchanged and remain highest in their nearby-cost groups, but Full is no longer the least-cost point in the higher group. The three ImageNet epochs take 1386.6, 1375.3, and 1360.7 seconds, so the short window does not establish long-window stationarity. Its $f{=}0.3$ control was stopped before completion and is excluded. Resource snapshots show no foreign GPU process on the probe devices. Headline costs keep the common 10-epoch audit basis, and these measurements are released as a separate validation.

\section{Robustness: Architecture and Training Budget}
\label{app:robustness}

We run two ablations on Tiny ImageNet, all 11 methods at fractions $\{0.1, 0.3, 0.5\}$ with 3 seeds each: (1) replacing ResNet-18 with ViT-Tiny trained for 300 epochs with a DeiT-style recipe (RandAugment, label smoothing, stochastic depth, AdamW with cosine schedule and warmup), and (2) reducing ResNet-18 training from 200 to 50 epochs. Table~\ref{tab:ablation-ranking} in Section~\ref{sec:tradeoff} reports the average rank of each method across (fraction, seed) groups with $95\%$ bootstrap intervals over seed resamples.

Rankings are consistent across the three settings: Spearman correlations against the main column are $\rho{=}0.94$ for short training and $\rho{=}0.78$ for ViT-Tiny (both $p{<}0.01$). Herding and Forgetting stay in the top tier, EL2N stays last at low fractions, and Random and Uniform stay competitive in the middle. The two noticeable shifts are Craig and GradMatch moving up under ViT-Tiny while kCenter drops to near-last. Absolute accuracies differ (ViT-Tiny trails ResNet-18 by $10$ to $17$\,pp, as expected for a patch-tokenized model trained from scratch on $64\times64$ images), while the method ordering is preserved.

\FloatBarrier

\section{Partial-Scan Pilot Details}
\label{app:partial-scan}

Class-stratified partial-scan Herding restricts both the selector training and the per-class greedy matching to a random $\rho$-fraction of each class, reducing the selection pipeline's data footprint to $\rho N$. On Tiny ImageNet with 10 seeds per cell, scanning only $\rho{=}0.5$ of the data preserves $92\%$ of full Herding's accuracy edge over Random at $f{=}0.05$ and $73\%$ at $f{=}0.10$, and Fig.~\ref{fig:partial-scan-fig} in Section~\ref{sec:tradeoff} plots the gain against $\rho$. The pilot is an accuracy proof of concept on Tiny ImageNet, with a CIFAR-10 replication of the cost pattern. The recorded selection times are ordinary run clocks, read only as a within-pilot comparison across $\rho$ on one host. Mean selection time on CIFAR-10 at $f{=}0.1$ is $477$\,s at $\rho{=}1$, $236$\,s at $\rho{=}0.5$, and $113$\,s at $\rho{=}0.2$, which is the same near-proportional pattern the main text reports for Tiny ImageNet. Welch's test against a 30-seed Random baseline gives $p{<}0.001$ in 7 of 8 settings and $p{=}0.083$ at $(f,\rho)=(0.05,0.2)$. Pool construction, scan schedule, and the interaction with other selectors are untested.

\section{Transfer to a Medical-Imaging Dataset: Retinal OCT}
\label{app:retinaloct}

We run the same protocol on Retinal OCT \citep{kermany2018identifying} (83K images, 4 classes, $224{\times}224$, ResNet-18, 200 epochs). All 11 methods are evaluated at fractions $\{0.05, 0.1, 0.3, 0.5, 0.7\}$ with 3 seeds each, except GradMatch at $f{\geq}0.5$ because OMP exceeds practical wall-clock budgets, for 159 subset-selection runs plus 3 Full-data baselines. The Full-data baseline reaches $99.97\%$.

The Retinal OCT ranking splits the same way as the natural-image grids. Coverage-preserving and simple methods are tightly bunched near the Full-data ceiling: at $f{=}0.05$, Moderate $99.76$, Herding $99.66$, Random $99.55$, Uniform $99.52$, Submodular $99.14$ (all within $1$\,pp of Full). Score-ranking methods that select hardest examples again collapse at low fractions: at $f{=}0.05$, EL2N $27.65\%$, GraNd $39.33\%$, GradMatch $53.48\%$, and at $f{=}0.10$ EL2N $48.73\%$, GraNd $60.64\%$, GradMatch $66.56\%$. Forgetting underperforms relative to its strong showing on the natural-image grid ($f{=}0.05$: $89.67\%$), reflecting that Retinal OCT has many easy correctly-classified images and few persistently-forgotten ones, weakening its signal. Ranking the 11 methods by mean accuracy over the fractions all of them share ($f{\le}0.3$), the Spearman correlation between the Retinal OCT ranking and each method's average rank over the natural-image datasets it runs on is $\rho{=}0.66$ ($p{=}0.03$), and the agreement is driven by the same split: score-ranking methods at the bottom, coverage-preserving and simple methods at the top. Retinal OCT is near saturation, which compresses the top of the ranking.

\end{document}